\documentclass{article}

\usepackage{arxiv}
\usepackage{graphicx}

\usepackage[utf8]{inputenc} 
\usepackage[T1]{fontenc}    
\usepackage{hyperref}       
\usepackage{url}            
\usepackage{booktabs}       
\usepackage{amsfonts}       
\usepackage{nicefrac}       
\usepackage{microtype}      
\usepackage{lipsum}

\newcommand\blfootnote[1]{%
  \begingroup
  \renewcommand\thefootnote{}\footnote{#1}%
  \addtocounter{footnote}{-1}%
  \endgroup
}

\title{JOINT CONTEXTUAL MODELING FOR ASR CORRECTION and LANGUAGE UNDERSTANDING}

\author{
  Yue Weng\thanks{Equal contribution. Order chosen randomly}
   \And Sai Sumanth Miryala\footnotemark[1] \And Chandra Khatri\footnotemark[1] \And Runze Wang \And Huaixiu Zheng \And Piero Molino
  \And Mahdi Namazifar \And Alexandros Papangelis \And Hugh Williams \And Franziska Bell 
  \AND 
  Gokhan Tur \\
 Uber AI \\
}

\begin{document}
\maketitle
\blfootnote{%
  \footnotesize \textcopyright Copyright 2020 IEEE. Published in the IEEE 2020 International Conference on Acoustics, Speech, and Signal Processing (ICASSP 2020), scheduled for 4-9 May, 2020, in Barcelona, Spain. Personal use of this material is permitted. However, permission to reprint/republish this material for advertising or promotional purposes or for creating new collective works for resale or redistribution to servers or lists, or to reuse any copyrighted component of this work in other works, must be obtained from the IEEE. Contact: Manager, Copyrights and Permissions / IEEE Service Center / 445 Hoes Lane / P.O. Box 1331 / Piscataway, NJ 08855-1331, USA. Telephone: + Intl. 908-562-3966.}
\newcommand\copyrightnotice{%
\begin{tikzpicture}[remember picture,overlay]
\node[anchor=south,yshift=10pt] at (current page.south) {\fbox{\parbox{\dimexpr\textwidth-\fboxsep-\fboxrule\relax}{\copyrighttext}}};
\end{tikzpicture}%
}

\begin{abstract}
The quality of automatic speech recognition (ASR) is critical to Dialogue Systems as ASR errors propagate to and directly impact downstream tasks such as language understanding (LU). In this paper, we propose multi-task neural approaches to perform contextual language correction on ASR outputs jointly with LU to improve the performance of both tasks simultaneously. To measure the effectiveness of this approach we used a public benchmark, the 2nd Dialogue State Tracking (DSTC2) corpus. As a baseline approach, we trained task specific Statistical Language Models (SLM) and fine-tuned state-of-the-art Generalized Pre-training (GPT) Language Model to re-rank the n-best ASR hypotheses, followed by a model to identify the dialog act and slots. i) We further trained ranker models using GPT and Hierarchical CNN-RNN models with discriminatory losses to detect the best output given n-best hypotheses. We extended these ranker models to first select the best ASR output and then identify the dialogue act and slots in an end to end fashion. ii) We also proposed a novel joint ASR error correction and LU model, a word confusion pointer network (WCN-Ptr) with multihead self attention on top, which consumes the word confusions populated from the n-best. We show that the error rates of off the shelf ASR and following LU systems can be reduced significantly by 14\% relative with joint models trained using small amounts of in-domain data.
\end{abstract}

\keywords{Language understanding, Multi-task learning, Generalized language models, ASR reranking
}

\section{Introduction}
Goal-oriented dialogue systems aim to automatically identify the intent of the user as expressed in natural language, extract associated arguments or slots, and take actions accordingly to satisfy the user’s requests~\cite{SLUBook}. In such systems, the speakers' utterances are typically recognized using an ASR system. Then the intent of the speaker and related slots are identified from the recognized word sequence using an LU component. Finally, a dialogue manager (DM) interacts with the user (not necessarily in natural language) and helps the user achieve the task that the system is designed to support. As a result, the quality of ASR systems has a direct impact on downstream tasks such as LU and DM. This becomes more evident in cases where a generic ASR is used, instead of a domain-specific one \cite{DBLP:conf/slt/MorbiniAASSGTN12}.


A standard approach to improve ASR output is to use 
an SLM or a neural model to re-rank different ASR hypotheses and use the one with the highest score for downstream tasks. 
Moreover, neural language correction models can also be trained to recover from the errors introduced by the ASR system via mapping ASR outputs to the ground-truth text in end-to-end speech recognition \cite[among others]{DBLP:conf/interspeech/TanakaMMA18}. In this paper we experiment with training ASR reranking/correction models jointly with LU tasks in an effort to improve both tasks simultaneously, towards End-to-End Spoken Language Understanding (SLU).



The major contributions of this work are as follows:
\begin{itemize}

\item[1.] Presented a cascaded approach to first select the best ASR output and then perform LU
\item[2.] Presented a novel alignment scheme to create a word confusion network from ASR n-best transcriptions to ensure consistency between model training and inference
\item[3.] Proposed a framework for using ASR n-best output to improve end-to-end SLU by multi-task learning, i.e. ASR correction, and LU (intent and slot detection). 
\item[4.] Proposed several novel architectures adopting GPT \cite{GPT2018} and Pointer network \cite{oriol:nips15} with a $2$D attention mechanism
\item[5.] Comprehensive experimentation to compare different model architectures, uncover their strengths and weaknesses and demonstrate the effectiveness of End-to-End learning of ASR ranking/correction and LU models.
\end{itemize}


\section{Related Work} \label{related_work}
\textbf{Word Confusion Networks:} A compact and normalized class of word lattices, called word
confusion networks (WCNs) 
were initially proposed for improving ASR performance ~\cite{lidia-es99}.
WCNs are much smaller than ASR lattices but have better or comparable word and oracle accuracy, and because of this they have been used for many tasks, including SLU~\cite[among others]{gokhan-icslp02}.
However, to the best of our knowledge they have not been used with Neural Semantic Parsers implemented by Recurrent Neural Networks (RNNs) or similar architectures. The closest work would be 
\cite{Ladhak2016LatticeRnnRN}, who propose to traverse an input lattice in topological order 
and use the RNN hidden state of the lattice final state as the dense vector representing the entire lattice. However, word confusion networks provide a much better and more efficient solution thanks to token alignments. We use this idea to first infer WCNs from ASR n-best and then directly use them for ASR correction and LU in joint fashion.

\textbf{ASR Correction:} Neural language correction models have been widely used to tackle a variety of tasks including grammar correction, text or spelling correction and completion of ASR systems. 
\cite{DBLP:conf/interspeech/TanakaMMA18, DBLP:journals/corr/abs-1902-07178} are highly relevant to our work as they performed spelling correction on top of ASR errors to improve the quality of speech recognition. However, our work differs significantly from existing work as we tackle neural language correction together with a downstream task (LU in this case) in a multi-task learning setting. In addition, we use the alignment information contained in the n-best list by an inferred word confusion network and input all n-best into a single neural network.

\textbf{Re-ranking and Joint Modeling:} 
\cite[among others]{DBLP:conf/cicling/MaCB17} 
showed that n-best re-ranking helps in reducing WER, while \cite{DBLP:conf/slt/MorbiniAASSGTN12, DBLP:conf/ijcnlp/CoronaTM17} showed that using ranking or in-domain language models or semantic parsers over n-best hypotheses significantly improves LU accuracy. Moreover, \cite{DBLP:conf/slt/Jonson06, DBLP:conf/interspeech/RajuHLGKMVR18} showcased the importance of context in ASR performance. However, none of the above-mentioned works involved joint or contextual modeling with end-to-end comparison. 
\cite[among others]{DBLP:conf/slt/HaghaniNBCGMPQW18} showcased that audio features can be directly used for LU, however, such systems are less robust for task completion, especially those which involve multi-turn state tracking. Moreover, another objective of our research is to evaluate if generalized language models such as GPT \cite{GPT2018} can be useful for joint ASR re-ranking and LU tasks.

\section{SLU Background and Baselines} \label{sec:slu_background}
\subsection{ASR Ranking and Error Correction}\label{sec:asr_ranking_correction}
To prevent the propagation of ASR errors to downstream applications such as NLU in a dialogue system, ASR error correction \cite{Roark2004CorrectiveLM, Kumar17} has been explored extensively using a variety of approaches such as language modeling and neural language correction. In the following, we cover the formulation of ASR error corrections using both approaches.

\textbf{Language Modeling}: Significant research has been conducted around count-based and neural LMs \cite[among others]{DBLP:conf/naacl/1993, DBLP:conf/icassp/2010}. 
Even though RNN-LMs have significantly advanced the state of the art (through re-ranking and Seq2Seq architectures), they still do not fully preserve the context, especially in ASR for Dialogue Systems, wherein context for a word might not correspond to words immediately observed before. Bidirectional and Attention based Neural LMs such as Embeddings for Language Models (ELMo) and Contextual Word Vectors (Cove) have shown some improvements \cite{DBLP:conf/naacl/PetersNIGCLZ18, DBLP:conf/nips/McCannBXS17}. 
More recently, Transformer Networks based LMs such as Bidirectional Encoder Representations from Transformers (BERT) \cite{DBLP:journals/corr/abs-1810-04805} and GPT \cite{DBLP:journals/corr/abs-1801-10198, GPT2018} have significantly outperformed most baselines in a variety of tasks.

\textbf{Statistical and Neural LMs for Re-ranking/Re-scoring:} We trained a variety of LMs on the DSTC2 training data, which are then used for re-ranking the ASR hypotheses based on perplexity. We trained the following LMs: (1) Count based word level Statistical Language Model (SLM) (experimented with several context sizes with backoff) (2) Transformer based OpenAI GPT LM \cite{GPT2018}, which uses Multi-headed Self-attention over the context followed by position-wise Feed-Forward layers to generate distribution over output sequence. While the GPT is trained with sub-word level LMs as proposed in the initial architecture. We start with a pre-trained GPT-LM released by OpenAI \cite{GPT2018} and then fine-tune on DSTC-2 data along with passing contextual information (past system and user turns along with current system turn separated by a special token) as input to the model. We experimented with the number of previous turns provided as context to the language model and picked the best configuration based on the development data. These LMs are used for re-ranking and obtaining the best hypothesis, which is then fed into a Bi-LSTM CRF \cite{huang2015bidirectional} for intent and slot detection, which are used as baselines.

\textbf{Neural Language Correction (NLC):} Neural language correction \cite{DBLP:journals/corr/XieAAJN16} aims at using neural architectures to map an input sentence $X=(x_1, \dots, x_{T_X})$ containing errors, to a ground-truth output sentence $Y=(y_1, \dots, y_{T_Y})$.
We use WCN (inferred from the n-best) to align the n-best list with the ground-truth. This way, the input $X$ and output $Y$ will have the same length 
and they are aligned at word-level: namely $x_i$ and $y_i$ are highly plausible pairs. As a result, we can use the same RNN decoder for slot tagging as described in Section \ref{sec:wcn}. Note that sequence tagging architectures can be used for multi-task learning with multiple prediction heads of word-correction and IOB tag prediction.

\subsection{Language Understanding}
The state-of-the-art in SLU relies on RNN or Transformer based approaches and its variations, which have first been used for slot filling by 
 \cite{yao2013RNN} and 
\cite{mesnil2013RNN} simultaneously. More formally, to estimate the sequence of tags $Y = y_1, ..., y_n$ in the form of IOB labels as in~\cite{raymond-riccardi07} (with 3 outputs corresponding to `B', `I' and `O'), and corresponding to an input sequence of tokens $X = x_1, ..., x_n$, the RNN architecture consists of an
input layer, a number of hidden layers, and an output layer. 
Nowadays, state-of-the-art slot filling methods usually rely on sequence models like RNNs~\cite[among others]{dilekIS16,RNN-TASL}.
Extensions include encoder-decoder models
~\cite{bingIS16,zhu2016}, transformers~\cite{DBLP:journals/corr/abs-1902-10909}, or memory~\cite{vivianIS16}. Historically, intent determination has been seen as a classification problem and slot filling as sequence classification problem, and in the pre-deep-learning era these two tasks were typically modeled separately. 
To this end ~\cite{dilekIS16} proposed a single RNN architecture that integrates intent detection and slot filling. 
The input of this RNN is the input sequence of words (e.g., user queries) and the output is the full semantic frame (intent and slots).

\section{Joint ASR Correction and NLU Models}
\label{modeling}
\subsection{Word Confusion Network and N-best Alignment}\label{sec:wcn}
N-best output from out of box ASR systems are usually not aligned. So, for WCN based models (Section~\ref{sec:approach-wcn-pointer}), an extra step is needed to align the n-best. 
Here's our approach: Use the word level Levenshtein distance to align every ASR hypothesis with the one-best hypothesis (as we do not have the transcription during testing). To unify these n-references, we merge insertions across all hypotheses to create a global reference \(R_{global}\), which is then used to expand all the original n-best to obtain hypotheses of same length as \(R_{global}\). During training, we align transcriptions with \(R_{global}\) for and NLU tasks such as tagging experiments. 

\begin{table*}[t]
\small	
\centering
\begin{tabular}{|c|c|c|c|c|c|c|} 
 \hline
Experiments & WER & SER & DA-Acc & Slot-F1 & TER	 & FER\\
\hline\hline
\textit{1-best (C)} & 29.99	 & 54.48	 & 88.28 & 81.01 & 14.9 & 21.43 \\
\textit{oracle (C)} &19.83	 &41.0	 &90.37	 &85.78
&9.14 & 17.69 \\
\textit{ground truth (C)} & 0	 & 0	 &94.91	 &98.88
&0.31 & 5.42 \\
\hline
\textit{SLM (C)} &	27.95 & 51.94 & 87.75 & 79.45	&14.83 & 22.17\\
\textit{GPT-LM (C)} &	26.82 &49.9	 &89.09	 &81.3	& 15.20& 21.07\\
\hline
\textit{Hier-CNN-RNN\_Ranker (C)}& 25.84 &\textbf{47.92}	 &88.67	 &81.94	&13.14 &20.3	\\
\textit{GPT\_Ranker (C)} &26.06	 &49.13	 &89.58	 &82.38	& 13.81&19.28 \\
\hline
\textit{WCN\_Pointer\_Head\_No\_Attention (J)} & 26.99 &49.13	 & 89.67	 &81.84	& 13.18& 20.87\\
\textit{WCN\_Pointer\_Head\_Multiheads\_Attention (J)} &26.73	 &49.01	 &89.48	 &82.04	&\textbf{13.06} &20.82 \\
&  & 	 & 90.39 (C) & 82.06 (C) & 13.05(C) & 20.79 (C) \\
\textit{WCN\_Word\_Generation\_Head\_Multiheads\_Attention (J)} &-	 &-	 & 89.48 (J)	 &\textbf{83.21 (J)}	& 13.51 (J)& \textbf{19.71} (J) \\ &  & 	 & 89.76 (C) & 82.68 (C) & 13.39(C) & \textbf{18.45(C)} \\
\hline

\textit{GPT\_MultiHead (J)} & 25.97 & 49.05	 & \textbf{92.69 (J)} & 77.8 (J)& 13.24 (J)& 23.77 (J) \\
 &  & 	 & 89.73 (C) & 82.78 (C) & 13.63 (C) & 19.09 (C) \\
\textit{GPT\_MultiHead\_Context (J)} & \textbf{25.80} & 48.65	 &	92.28 (J) & 78.09 (J) & 13.08 (J) & 23.09 (J)\\
& & 	 &	89.82 (C) & 82.67 (C) & 13.56 (C) & 18.92 (C) \\
 \hline
\end{tabular}
    \caption{\textit{SLU results. All the ``J" models perform LU in joint manner, while  for the other models (C), output of the ranker or correction module is fed into a separate BiLSTM-CRF for tagging and dialogue act detection. The ASR correction performance is masked to ensure fair comparison for WCN-Word-Generation-Head-MultiHeads-Attention.}}
\label{tab:results}
\end{table*}

\subsection{GPT based Joint SLU}\label{sec:gpt_model}
\begin{figure}
  \centering
  \includegraphics[scale=0.2]{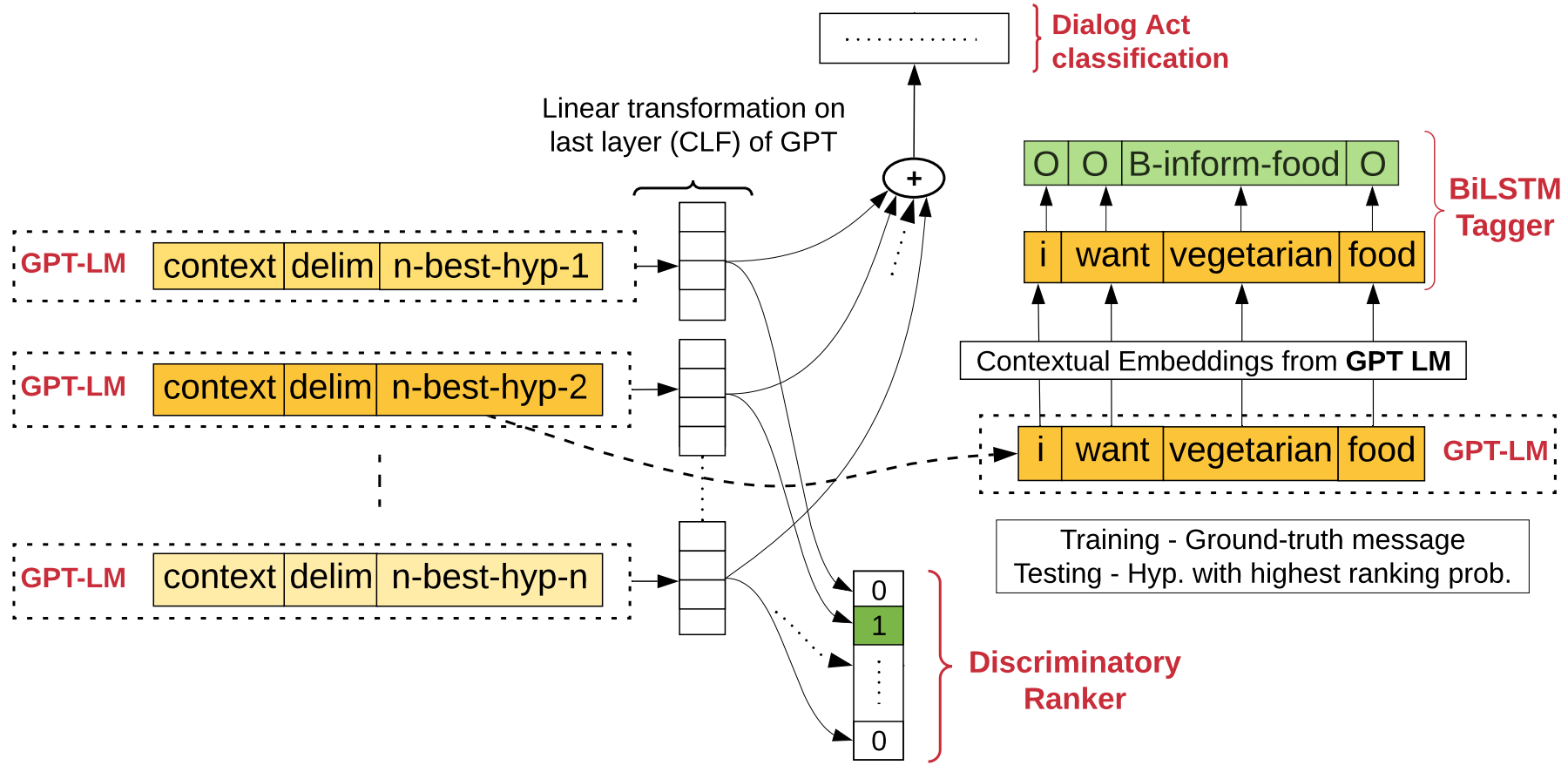} 
  \caption{\textit{GPT joint model for ranking, intent and slot detection}}
  \label{fig:gpt_arch}
\end{figure}

As described in Section \ref{sec:slu_background}, GPT based LM is used for re-scoring the n-best hypothesis. We extend the GPT-LM with three additional heads (Figure \ref{fig:gpt_arch}): Discriminatory Ranking, Dialogue Act Classification, and Slot Tagging. In addition to the likelihood of the sequence obtained from the LM, we train a discriminatory ranker  to select the oracle. 

The ranker takes the last state (or `clf' token embedding) as input for each hypothesis and outputs 1 if it is oracle or 0 otherwise. Similarly, we sum the last state for all the hypotheses and use it for Dialogue Act classification. For tagging, we use the transcription during training and hypothesis selected by the ranker during testing or validation. We add a Bi-LSTM layer on top of the embeddings obtained from GPT-LM to predict IOB tags. The model inputs are context (last system and user utterance, current system utterance) and n-best hypothesis, all separated by a delimiter used in the original GPT. 

\subsection{Hierarchical CNN-RNN Neural Ranker}\label{sec:Hier-CNN-RNN}

Given the n-best as input, we built a multi-head hierarchical CNN-RNN (Hier-CNN-RNN) model to predict the index of the oracle directly. The nbest ASR hypothesis is first input to a 1D Convolutional Neural Network (CNN) to extract the n-gram information. The motivation to use CNN is to align the words in the n-best hypothesis since the convolutional filters are  invariant to translation. The features extracted from CNN are then fed to a RNN to capture the sequential information in the sentences. The hidden states from RNN are concatenated together.
The last hidden states from all n-best is averaged to predict the index of the oracle in n-best. For the joint model, the predicted oracle is fed into a LU head module to predict the intent and slots. The joint model did not perform well, so we have excluded it from the results in the interest of space. 

\subsection{WCN Pointer Joint Neural Correction and NLU}
\label{sec:approach-wcn-pointer}

\begin{figure}
   \centering
   \includegraphics[scale=0.095]{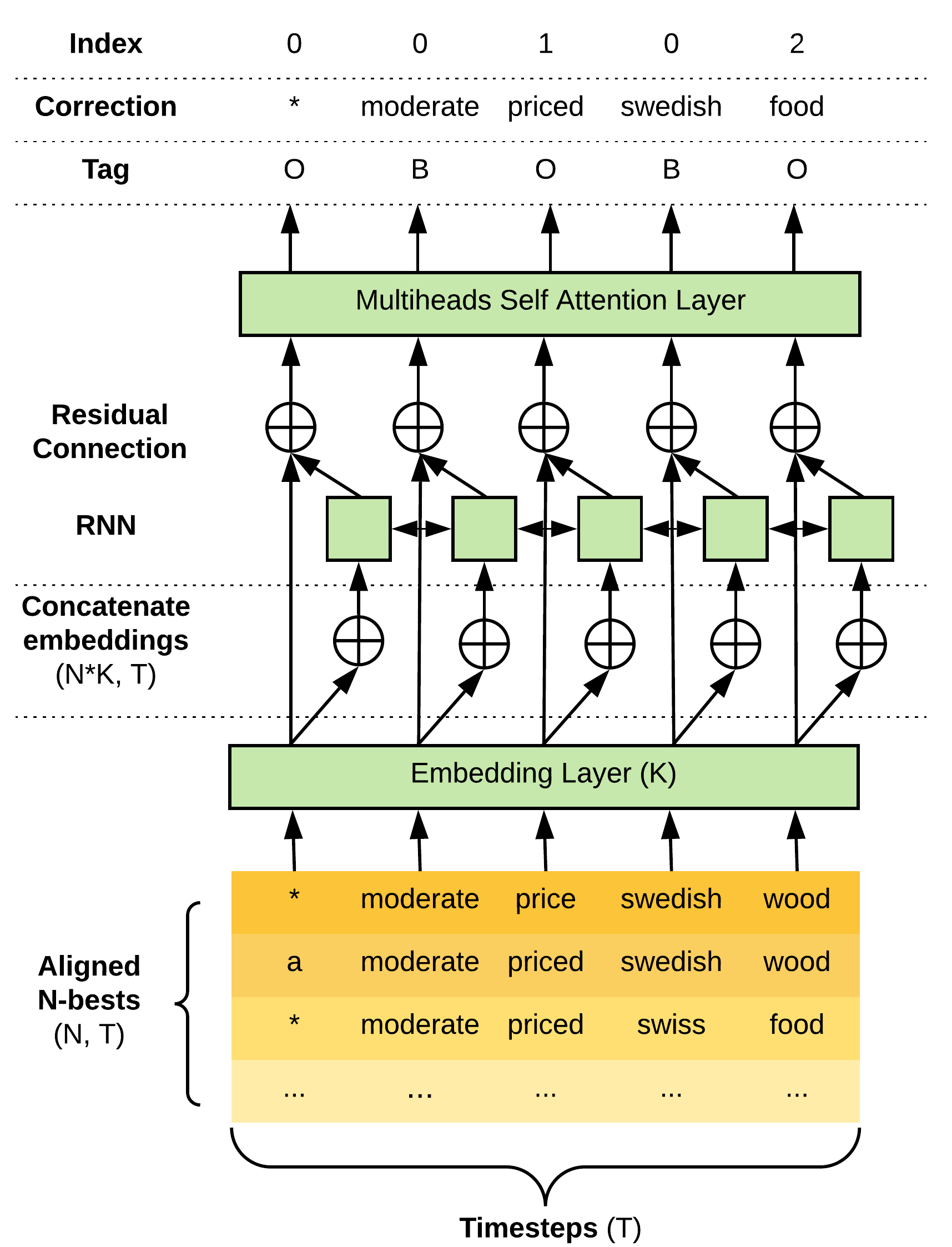}
    \caption{\textit{WCN Pointer Model With Multiheads Self Attention}}
    \label{fig_wcn_arch}
\end{figure}

The WCN model, as illustrated in Figure \ref{fig_wcn_arch}, takes all the N-best in at the same time. Specifically, for a given n-best, a word confusion network alignment is constructed. Then, for each time step, the model concatenates the embeddings of all its n-best into a word bin and processes them through a multi-headed Bi-LSTM, where each hidden states is concatenated with embedding vectors as residual connection. Next, a multihead self attention layer is applied to all hidden states, which, in addition to predicting the IOB tags, generates the correct word based on vocabulary (word generation head) or predicts the index at which the correct word is found (pointer head) for each time step. If there is no correct words, we select first best. We append an EOS token in the last time step and use the last hidden state for intent prediction. The rationale behind this is that the correct word often exists in the WCN alignment but can be at different positions. 

\section{Experiments and Results}
\label{XP_results}

\textbf{Data:} We use DSTC-2 data~\cite{dstc2}, wherein the system provides information about restaurants that fit users' preferences (price, food type, and area) and for each user utterance the 10-best hypotheses are given. 
We modified the original labels for dialogue acts to a combination of dialogue act and slot type (e.g. dialogue act for ``whats the price range" becomes ``request\_pricerange" instead of ``request"), which gets us a total of 25 unique dialogue acts instead of initial the 14. Further, we address the slot detection problem as a slot tagging problem by using the slot annotations and converting them into IOB format. In our analysis, we ignore the cases that have empty n-best hypotheses or dialogue acts, and those with the following transcriptions: ``noise", ``unintelligible", ``silence",  ``system", ``inaudible", and ``hello and welcome". This leads to 10,881 train, 9,159 test, and 3,560 development utterances. Our objective is not to out-perform the state-of-the-art approaches on DSTC-2 data, but to evaluate if we can leverage ASR n-best in a contextual manner for overall better LU through multi-task learning. 
We also plan to release the data for enabling future research. 

\textbf{Baseline and Upper Bound:} We obtain WER and sentence error rate (SER) to evaluate ASR and dialogue act accuracy (DA-Acc), tag error rate (TER), slot F1, and frame error rate (FER) to evaluate LU. 
We compare the metrics obtained for joint models with the ones through cascading (i.e. non-joint models).  For ASR, we consider three baselines: 1-best, SLM and GPT based re-ranked hypothesis. For LU, we trained a separate Bi-LSTM CRF tagger with an extra head for Dialogue Act classification, which we run on top of the three baselines mentioned above to obtain LU baseline numbers. To better understand the upper-bound, we obtain the metrics for the oracle and ground truth transcription as well.

\subsection{Results and Discussion}
As shown in Table \ref{tab:results}, it can be observed that all models outperform the 1-best in ASR metrics. Even SLM trained and GPT-LM fine-tuned on 11k training utterances perform significantly better than the 1-best on ASR metrics. However this does not translate into improvement in the LU metrics. In fact, the output reranked using SLM does worse on the LU metrics. This indicates that just reducing WER and SER doesn't lead to improvement in LU. 
The Hier-CNN-RNN Ranker model achieves 14\% lower WER while also improving the LU metrics (5.2\% reduction in FER). The GPT based discriminatory ranker also improves both ASR (13\% reduction in WER) and LU (10\% reduction in FER). This indicates that training a discriminatory ranker which identifies the oracle would do better than training a task-specific Language Model. Some of the models even out-perform the oracle on DA-Acc ($>$2\% absolute improvement) because the Dialogue Act prediction head uses an encoding of all hypotheses (including oracle). 

On the other hand, WCN models lead to the best LU slot tagging performance. WCN models out-perform the baseline with 2.2\% absolute improvement in slot F1 score, 12\% TER reduction and most importantly 8\% FER reduction. The GPT joint models on the other hand improve the TER but their slot F1 is significantly lower compared to the GPT ranker. This is probably because there are a lot more `O' tags compared to `B' and `I'. We noticed that we were able to achieve even higher accuracy by running the baseline tagger on the corrected output of the joint models. Our lowest FER is achieved by running the baseline tagger model on the joint WCN model (with word generation head) output. While the WCN model's performance is improved by using the baseline tagger, the difference is much more profound for the GPT models (the frame error rate drops by almost 4\%). We believe this is because the WCN models consume aligned n-best, which improves the model learning efficiency and 
they converge better when data size is relatively small. Furthermore, we observed that adding multihead attention layer and multiple heads helps the WCN models across all metrics.     

\section{Conclusions}

We have presented a joint ASR reranker and LU model and showed experimental results with significant improvements on the DSTC-2 corpus. To the best of our knowledge this is the first deep learning based study to this end. We have also contrasted these models with cascaded approaches building state-of-the-art GPT based rankers. Our future work involves extending such end to end LU approaches towards tighter integration with a generic ASR model.

\bibliographystyle{unsrt}  
\bibliography{references}  

\end{document}